\def\year{2020}\relax
\documentclass[letterpaper]{article} 
\usepackage{aaai20}  
\usepackage{times}  
\usepackage{helvet} 
\usepackage{courier}  
\usepackage[hyphens]{url}  
\usepackage{graphicx} 
\usepackage{graphicx}  
\usepackage{amssymb}
\usepackage{amsmath}
\title{Show, Recall, and Tell: Image Captioning with Recall Mechanism}
\author{Li Wang\textsuperscript{\rm 1}\textsuperscript{,\rm 3}\thanks{Equal Contribution.},
         Zechen Bai\textsuperscript{\rm 2}\textsuperscript{,\rm 3}${^*}$,
        Yonghua Zhang\textsuperscript{\rm 3},
         Hongtao Lu\textsuperscript{\rm 1}\thanks{Corresponding author} \\
\textsuperscript{\rm 1}Department of Computer Science and Engineering, Shanghai Jiao Tong University, China\\
\textsuperscript{\rm 2}University of Sciences and Technology Beijing, China\\
\textsuperscript{\rm 3}AI-Lab Visual Search Team, Bytedance\\
liwang.sjtu@gmail.com, ustbbzch@gmail.com, mpeg21@hotmail.com, htlu@sjtu.edu.cn
}
 
\begin{document}

\maketitle

\begin{abstract}
Generating natural and accurate descriptions in image captioning has always been a challenge. In this paper, we propose a novel recall mechanism to imitate the way human conduct captioning. There are three parts in our recall mechanism : recall unit, semantic guide (SG) and recalled-word slot (RWS). Recall unit is a text-retrieval module designed to retrieve recalled words for images. SG and RWS are designed for the best use of recalled words. SG branch can generate a recalled context, which can guide the process of generating caption. RWS branch is responsible for copying recalled words to the caption. Inspired by pointing mechanism in text summarization, we adopt a soft switch to balance the generated-word probabilities between SG and RWS. In the CIDEr optimization step, we also introduce an individual recalled-word reward (WR) to boost training. Our proposed methods (SG+RWS+WR) achieve BLEU-4 / CIDEr / SPICE scores of 36.6 / 116.9 / 21.3 with cross-entropy loss and 38.7 / 129.1 / 22.4 with CIDEr optimization on MSCOCO Karpathy test split, which surpass the results of other state-of-the-art methods.
\end{abstract}

\section{Introduction}
Image captioning is defined as automatically generating a descriptive statement from an image. This task needs to exploit image information and then to generate a natural caption. Image captioning can be applied to a wide range of domains, for example, automatically adding subtitles to images or videos, which can do great help in search task.

\begin{figure}
    \centering
    \includegraphics[width=0.95\columnwidth]{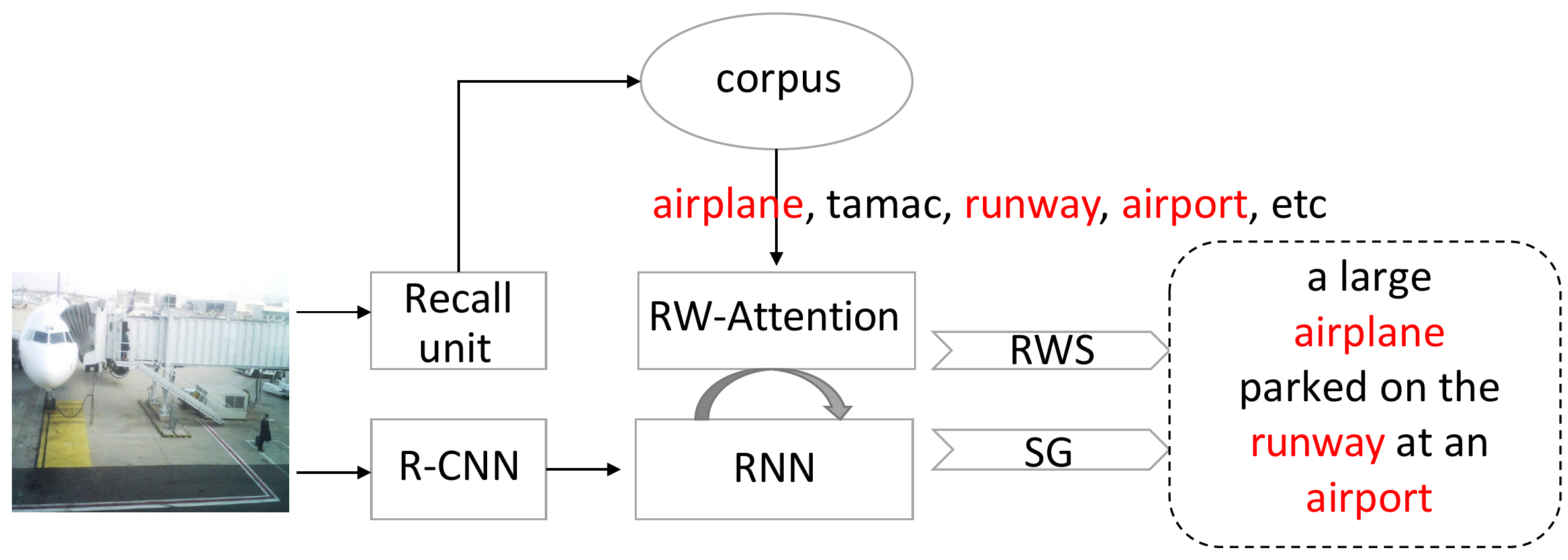}
    \caption{An overview of our proposed methods. Recall unit is a text-retrieval module. RW-Attention denotes recalled-words attention. RWS represents recalled-words slot. SG is semantic guide. In short, we introduce a recall unit to the traditional captioning model, employing recalled words to boost the performance of captioning.}
    \label{fig:1}
\end{figure}

In the last few years, encoder-decoder models have been designed to accomplish the captioning task in many methods \cite{socher2014grounded,vinyals2015show,qin2019look}. The role of encoder in captioning is to extract sufficient and useful visuqinal features from the image, and image has been mostly encoded by using Convolutional Neural Network (CNNs) such as ResNet \cite{He_2016_CVPR}. Meanwhile, the role of decoder is to exploit semantic part from encoded visual information and then decode it word by word. Recurrent neural network (RNNs) is the most commonly used method of decoder in captioning.

On one hand, visual attention methods \cite{xu2015show,lu2017knowing} have brought significant improvement in captioning on the most evaluation metrics like BLEU \cite{papineni2002bleu}, METEOR \cite{denkowski-lavie-2014-meteor} and CIDEr \cite{vedantam2015cider}. In the process of generating caption, visual attention methods can allocate different weights to different regions of an image, which prompt model only focus on the crucial parts of the image.

On the other hand, semantic methods have also improved captioning performance remarkably. Numerous methods \cite{You_2016_CVPR,lu2018neural} employ detection technique to obtain words of objects and attributes, then take those words as known objects to generate caption, but these methods are highly dependent on the performance of detection model. \cite{Mun2016TextguidedAM} have retrieved relevant guide texts according to the visual similarity between images, but they fail to construct a direct relationship between the guide texts and the generated caption.

Looking back the above methods of captioning, all information to generate caption is extracted from only one given image, but when a human describes an image, is all the information used only from this image? Usually not. When people are intend to do something, they will first recall past experiences, imitate them appropriately, and then do it. This is human instinct with no exception in captioning. People will recall how similar images were described, and then use these similar patterns to generate a caption for the image.

For the purpose of making captioning model to describe images in the way like human beings do, in this paper, we introduce a novel recall mechanism into captioning model. In order to recall useful and relevant words for each image, we apply an image-text matching model similar to that brought by \cite{lee2018stacked} as our text-retrieval model, and captions from training data are taken as our corpus. In this image-text matching model, we embed image feature and text feature into a common space, then calculate the cosine similarity between them. Triplet loss \cite{Kiros2014UnifyingVE,socher2014grounded} is the objective function for each mini-batch during training.

To make recalled words more relevant to the image and to filter out useless words, for each image, we construct a set of recalled words just from top K captions of text-retrieval task. As illustrated in Figure \ref{fig:1}, RW-Attention module is applied to obtain weights of recalled words, and these weights are used into two branches: semantic guide (SG), recalled-word slot (RWS), then the final caption is generated by these two branches.
As above, we mainly have the following contributions in this paper:
\begin{enumerate}
    \item In order to imitate the human behavior of recalling, we apply an image-text matching model to retrieve recalled words for each image.
    \item We propose two methods to utilize recalled words: semantic guide and recalled-word slot.
    \item In the CIDEr optimization stage, we propose a novel recalled-word reward to boost caption performance.
\end{enumerate}

We evaluate the performance of our proposed methods on MSCOCO Karpathy test split with both cross-entropy loss and CIDEr optimization. In order to fairly compare and convincingly prove the effectiveness of our methods, we incorporate our proposed methods into Up-Down model \cite{anderson2018bottom} and take it as our baseline model. It is shown that our approaches have obtained remarkable improvement over our baseline model. Our methods achieve BLEU-4 / METEOR / ROUGE-L / CIDEr / SPICE scores of 36.6 / 28.0 / 56.9 / 116.9 / 21.3 with cross-entropy loss and 38.5 / 28.7 / 58.4 / 129.1 / 22.4 with CIDEr optimization. We also conducted a series of experiments, taking several state-of-the-art models as baseline model and introducing our proposed methods into them respectively, which confirmed the effectiveness and generality.

\section{Related Work}
\textbf{Image captioning}. Most modern computer vision methods \cite{socher2014grounded,karpathy2015deep} encode image through CNNs and then decode it with RNNs. \cite{vinyals2015show} firstly incorporated attention mechanism into captioning. In this way, decoder can better extract local information from image, thus visual features can be better represented. Adaptive attention \cite{lu2017knowing} introduced a sentinel gate mechanism into visual attention, which can prompt the extent model focuses on visual features or semantic context. \cite{yao2017boosting} added the attributes information to the captioning model, which has greatly improved the performance of object description in captioning. SCA-CNN \cite{chen2017sca} has introduced channel-wise attention into captioning model. By this way, visual features can be better gathered by focusing on crucial channels. \cite{anderson2018bottom} have employed Faster R-CNN network pre-trained on Visual Genome  \cite{krishna2017visual} to generate more explicit features. Several region features with high confidence gathered as visual feature, which has shown remarkable advantage over CNN feature. \cite{Rennie_2017_CVPR} have applied reinforcement learning method to captioning. By this way, the model can be optimized directly on those objective evaluation metrics like CIDEr \cite{vedantam2015cider}, BLEU \cite{papineni2002bleu} and etc. The consistency between training objective and evaluation metric has improved caption performance on evaluation scores. \cite {luo2018discriminability} have incorporated a discriminative loss in reinforcement learning step, which has enhanced the diversity of caption.\\
\textbf{Image-text matching}. Image-text matching is used to evaluate the relevance between image and text, so we employ image-text matching model to accomplish our text-retrieval task. There have been numerous studies exploring encoding whole image and full sentences to common semantic space for image-text matching.
Learning cross-view representations with a hinge-based triplet ranking loss was first attempted by \cite{kiros2014unifying}. Images and sentences are encoded by deep Convolutional Neural Networks (CNNs) and Recurrent Neural Networks (RNNs) respectively. \cite{faghri2017vse++} addressed hard negative cases in the triplet loss function and achieve notable improvement. \cite{gu2018look} proposed a method integrating generative objectives with the cross-view feature embedding learning. \cite{lee2018stacked} suggested the alignment between objects or other stuffs in images and the corresponding words in sentences.\\
\textbf{Pointing mechanism}. Inspired by pointing mechanism \cite{see2017get}, an adaptive and soft switch is applied in this paper to accommodate the word probabilities between generate mode (SG) and copy mode (RWS). In this way, captioning model can switch freely between SG and RWS. 

\begin{figure*}[h]
    \centering
    \includegraphics[width=0.95\textwidth]{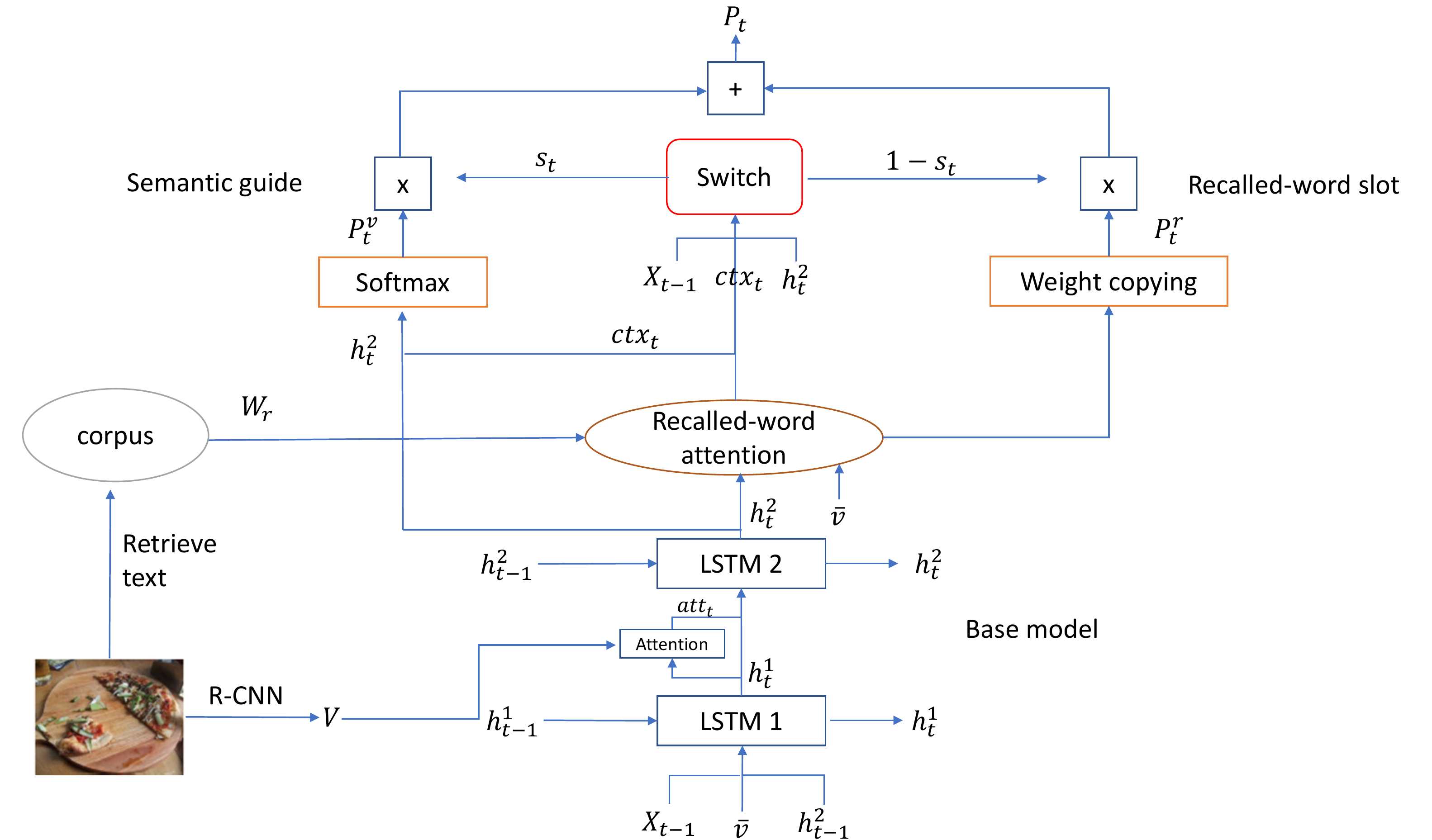}
    \caption{An illustration of our proposed recall mechanism. There are 3 parts in this figure: base model, semantic guide (SG), and recalled-word slot (RWS). We incorporate our recall mechanism into the base model Up-Down \cite{anderson2018bottom}. SG and RWS compute word probability individually, then we employ a soft switch to combine them into a final word probability $P_{t}$.}
    \label{fig:2}
\end{figure*}

\section{Methods} 
In this section, we will present our proposed methods in detail. For gaining better performance of caption, as in \cite{anderson2018bottom}, we extract the R-CNN feature for image: $V=\{v_{1},v_{2},...,v_{k}\}, v_{i} \in \mathbb{R}^{D}$.  We take the mean pooling vector $\bar{v}$ as the global visual feature.
\subsection{Text-Retrieval module}
For a given image $I$, the mean pooling vector $\bar{v}$ is taken as the visual feature. We embed it by a fully connected layer $W_{I}$:
\begin{equation}
    f(I)=W_{I}\bar{v}
\end{equation}

For a caption $C$, we utilize word2vec \cite{mikolov2013efficient} to embed words in $C$: 
\begin{equation}
            c_{i}={\rm word2vec}(w_{i})
\end{equation}

Where $c_{i}$ is the embedding of word $ w_{i}$, we denote $\hat{C} = \{c_{1},...,c_{n}\}$, where $n$ is the number of words in the caption.
Then we feed those words embedding into RNN:
\begin{equation}
            S={\rm RNN}(\hat{C})
\end{equation}

The output vectors of RNN are $S = \{s_{1},...,s_{n}\}$, where $s_{i}$ represents $i$-th word feature through RNN. Then we apply attention mechanism to predict the weights of word features with respect to image feature $f(I)$. For each vector $s_{i}$ in $S$, the weight is calculated as following:
\begin{align}
u_{i,t} &= w_{s}tanh(W_{s,u}s_{i}+W_{v}\bar{v})\\
\alpha_{t} &= softmax(u_{t})
\end{align}

Where $w_{s}$ , $W_{s,u}$ and $W_{v}$ are learned parameters for attention part, and $\alpha_{t} $ are the weights of word features. Then we take the weighted sum of word features as the feature for caption $C$:
\begin{align}
g(C) &= \sum_{i=1}^{n}\alpha_{i,t}s_{i}  
\end{align}

The similarity score between image $I$ and caption $C$ is computed as the cosine similarity:
\begin{equation}
s(I,C) = \frac{f(I)\cdot g(C)}{\left \| f(I)  \right \|\left \|g(C)\right \|}
\end{equation}

The triplet loss is the ranking objective for our text-retrieval model, we employ a hard negative hinge-based triplet loss with margin $\alpha$ as in \cite{lee2018stacked}:
\begin{align}\label{tri}
\begin{split}
L_{tri}(C,I)&= \max_{\hat{C}}[\alpha + s(I,\hat{C})-s(I,C)]_{+} \\ &+ \max_{\hat{I}}[\alpha + s(\hat{I},{C})-s(I,C)]_{+}
\end{split}
\end{align}

Where $[x]_{+}\equiv max(x,0)$. For a mini-batch, pairs $(I,C)$ are positive pairs, and pairs $(\hat{I},C)$ and pairs $(I,\hat{C})$ are negative pairs. The hard negative triplet loss in Eq.(\ref{tri}) tries to make positive pairs $(I,C)$ score higher than the maximum score of negative pairs with a margin $\alpha$.

\subsection{Captioning model}
As shown in Figure \ref{fig:2}, there are 3 parts of our captioning model: base model, semantic guide, and recalled-word slot. 
\subsubsection*{Base model}
We take Up-Down model \cite{anderson2018bottom} as our base model since its remarkable performance. Our proposed recall mechanism is implemented on it. So we will briefly introduce Up-Down model. As illustrated in Figure \ref{fig:2}, for each step, the input of LSTM1 is a concatenated vector of last word embedding $X_{t-1}$, the global pooling vector $\bar{v}$, and the last output $h_{t-1}^{2}$ from LSTM2. We attend the hidden output vector $h_{t}^{1}$ of LSTM1 to visual feature $V$, then feed the concatenated vector of visual context $att_{t}$ and $h_{t}^{1}$ into LSTM2. We denote [.] as the concatenation operation. The detailed formulas are explained as following:
\begin{align}
    h_{t}^{1} &= {\rm LSTM1}([X_{t-1},\bar{v},h_{t-1}^{2}])\\
    g_{i,t} &= w_{v}tanh(W_{v}^{1}v_{i}+W_{h}^{1}h_{t}^{1})\\
    \alpha_{t}^{v} &= softmax(g_{t})  \\
    att_{t} &= \sum_{i=1}^{k}\alpha_{i,t}^{v}v_{i}\\
    h_{t}^{2} &={\rm LSTM2}([att_{t},h_{t}^{1}])
\end{align}

Where $W_{v}^{1}$, $W_{h}^{1}$ and $w_{v}$ are learned parameters for visual attention part. $\alpha_
{t}^{v} = \{\alpha_{1,t},\alpha_{2,t},...,\alpha_{k,t}\} \in \mathbb{R}^{k} $ are the relevant weights of $V$, which sum to 1. $att_{t}$ is the weighted sum of $V$, which represents the most relevant locations of words to generate.

\subsubsection*{Semantic guide and recalled-word slot}
By applying our text-retrieval module to corpus, certain recalled words are collected for each image: $W_{r} = \{w_{r,1},w_{r,2},..w_{r,m}\}$, $m$ is the number of recalled words for an image. $X_{r} = \{x_{r,1},x_{r,2},..x_{r,m}\}$ is the corresponding embedding for recalled words.

As illustrated in Figure \ref{fig:2}, firstly, we apply attention mechanism  to obtain the weights of recalled words. The reason why we choose the concatenated vector $[h_{t}^{2},\bar{v}]$ to attend $X_{r}$ is we regard $[h_{t}^{2},\bar{v}]$ as an unit of the  semantic and visual information, which can be an accurate guide. The weights of recalled words are calculated as following:
\begin{align}
    r_{i,t} &= w_{x}tanh(W_{r}x_{i}+W_{h}^{2}h_{t}^{2}+W_{v}^{2}\bar{v})\\
    \alpha_{t}^{r} &= softmax(r_{t})
\end{align}   

Where $W_{r}$, $W_{v}^{2}$, $W_{h}^{2}$ and $w_{x}$ are learned parameters in recalled-word attention module, $\alpha_
{t}^{r} = \{\alpha_{1,t},\alpha_{2,t},...,\alpha_{m,t}\} \in \mathbb{R}^{m} $ represents the weights in recalled words.

We then apply the weights $\alpha_{t}^{r}$ into two branches: semantic guide, and recalled-word slot.
In the semantic guide branch, we obtain the recalled content $ctx_{t}$ by the weighted sum of recalled-word embedding, which can help to generate word probability distribution $P_{t}^{v}$:
\begin{align}
ctx_{t} &= \sum_{i=1}^{m}\alpha_{i,t}^{r}x_{i}\\
P_{t}^{v} &= softmax(W_{l} [ctx_{t},h_{t}^{2}])
\end{align}

Where $W_{l}$ is a logit layer to predict word probability $P_{t}^{v}(w)$ by softmax function.

In recalled-word slot,  we introduce a weight copying layer to copy the weights from recalled words directly to the word probability distribution $P_{t}^{r}$: 
\begin{equation}
P_{t}^{r}(w)=   
\begin{cases} 
\alpha_{t}^{r}(w) &  w \in W_{r}\\
0 & w \notin W_{r}
\end{cases}
\end{equation}

Where $\alpha_{t}^{r}(w)$ is the weight of word $w$ in $W_{r}$. If word $w$ does not exist in $W_{r}$, we set 0 to its probability. $P_{t}^{r}(w)$ represents the word probability from RWS branch. Since $P_{t}^{r}(w)$ only keeps the probability of recalled words, it seems to build a slot from recalled words to output caption.

We compute the final word probability by integrating two probabilities by a soft switch:

\begin{align}
\begin{split}
    s_{t} = &\sigma(W_{s,h}h_{t}^{2} + W_{s,c}ctx_{t}+ \\
    &W_{s,x}X_{t-1} + b_{s})
\end{split}\\
    P_{t}(w) &= (1-s_{t})P_{t}^{v}(w) + s_{t}P_{t}^{r}(w)
\end{align}

Where $\sigma(.)$ is a sigmoid function. $W_{s,h}, W_{s,c}, W_{s,x}$ and bias $b_{s}$ are the learned parameters to compute $s_{t} \in [0,1]$, which is considered as a soft switch between semantic guide and recalled-word slot. $P_{t}(w)$ is the weighted-sum probability of $P_{t}^{v}(w)$ and $P_{t}^{r}(w)$.

Take a overall look at our proposed methods. In semantic guide branch, $ctx_{t}$ is considered as a recalled content, merged with hidden vector $h_{t}^{2}$ to generate words.  In recalled-word slot, words generation are conducted by copying weights directly. Therefore, the semantic guide branch generates word by ``deep consideration'' with visual and recalled information, and the recalled-word slot is more like ``intuition''. The soft switch $s_{t}$ plays a role in combining them together to choose the most reasonable words.

\subsection{Objective}
Given an image $I$, a target ground truth sequence $w_{1:T}^{*}$. and an image captioning model with parameters $\theta$, we minimize the cross entropy loss as following:
\begin{align}
\begin{split}
L_{mle}(\theta) = &-\sum_{t=1}^{T}{\rm log}(p_{\theta}(w_{t}^{*}|w_{1:t-1}^{*}))\\
                = &-\sum_{t=1}^{T}{\rm log}(P_{t}(w_{t}^{*}|w_{1:t-1}^{*}))\\
                   = &-\sum_{t=1}^{T}{\rm log}\left ( \right.(1-s_{t})P_{t}^{v}(w_{t}^{*}|w_{1:t-1}^{*}) \\
                   &+s_{t}P_{t}^{r}(w_{t}^{*}|w_{1:t-1}^{*}) \left.\right)
\end{split}
\end{align}

Where $w_{t}^{*}$ is the word from the ground truth sequence at step $t$. In order to boost performance of captioning model, and to compare with recent work \cite{anderson2018bottom,Rennie_2017_CVPR}, we also apply CIDEr optimization to our training process. Initializing from the cross entropy model, the traditional CIDEr optimization \cite{Rennie_2017_CVPR} approach focuses on optimizing the CIDEr scores of the generated sentences. The training process is to minimize negative expected reward:
\begin{equation}
    L_{r}(\theta) = -\mathbb{E}_{w_{1:T}\sim p_{\theta}}[r(w_{1:T})]
\end{equation}

Where $r$ is the CIDEr function that can give a score for a sequence $w_{1:T}$. Following the method in SCST \cite{Rennie_2017_CVPR}, we approximate the gradient as:
\begin{equation}
\begin{split}
    \nabla_{\theta}L_{r}(\theta)\approx&-(r(w_{1:T}^{s})-r(w_{1:T}^{g}))\times\\
    & \nabla_{\theta}{\rm log}(p_{\theta}(w_{1:T}^{s}))
\end{split}
\end{equation}

Where $w_{1:T}^{s}$ is the sampled caption from the final probability distribution $P_{t}(w)$, and $w_{1:T}^{g}$ represents the caption obtained by greedily decoding. We take the score $r(w_{1:T}^{g})$ as the baseline, which is used to reduce the variance in training process.

\subsubsection*{Recalled-word reward}
Due to the fact that there are two branch probabilities in our model, we make a change to boost traditional CIDEr optimization. We propose an individual reward for recalled-word slot:
\begin{equation}
    r(W_{r}) = r(w_{1:T}^{s}) - r(w_{1:T}^{\hat{s}}) 
\end{equation}

Where $w_{1:T}^{s}$ is a sampled caption from the final probability distribution $P_{t}(w)$, and  $w_{1:T}^{\hat{s}}$ represents the sampled caption with soft switch $s_{t} = 0$ at all steps in generation. $s_{t} = 0$ indicates that we cut off recalled-word slot and only sample caption from $P_{t}^{v}(w)$.  In this way, recalled-word reward  $r(W_{r})$ can certify how much the improvement is from recalled-word slot. We minimize negative expected reward as following:
\begin{equation}\label{equ:lam}
\begin{split}
    L_{r}(\theta) = &-\lambda \mathbb{E}_{w_{1:T}^{\hat{s}}\sim p_{\theta}^{v}}[r(w_{1:T}^{\hat{s}})]\\ &-(1-\lambda) \mathbb{E}_{w_{1:T}^{s}\sim p_{\theta}}[r(W_{r})]
\end{split}
\end{equation}

\begin{equation}
\begin{split}
     \nabla_{\theta}L_{r}(\theta)\approx&-\lambda(r(w_{1:T}^{\hat{s}})-r(w_{1:T}^{\hat{g}}))\times\\
     &\nabla_{\theta}{\rm log}(p_{\theta}^{v}(w_{1:T}^{\hat{s}}))-\\
     &(1-\lambda)r(W_{r})\nabla_{\theta}{\rm log}(p_{\theta}(w_{1:T}^{s})) 
\end{split}
\end{equation}

Where $w_{1:T}^{\hat{g}}$ represents the caption obtained by greedily decoding by cutting off recalled-word slot, which is viewed as the baseline for $w_{1:T}^{\hat{s}}$. We do not introduce a baseline for $r(W_{r})$, because it is originally the result of subtraction, so the variance is relatively small.\\

\section{Experiments and results}
\subsection{Datasets}
\subsubsection*{MSCOCO}
We use the MSCOCO 2014 captions dataset \cite{lin2014microsoft} to evaluate our proposed method. As the largest English image caption dataset, MSCOCO contains 164,062 images. In this paper, we employ the ‘Karpathy’ splits \cite{karpathy2015deep} for validation of model hyperparameters and offline evaluation. This split has been widely used in prior works, choosing 113,287 images with five captions each for training and 5000 respectively for validation and test. For quantitative performance evaluation, we use the standard automatic evaluation metrics, namely SPICE \cite{anderson2016spice}, CIDEr \cite{vedantam2015cider}, METEOR \cite{denkowski-lavie-2014-meteor}, ROUGE-L  \cite{lin2004rouge} and BLEU \cite{papineni2002bleu}

\subsubsection*{Visual Genome}
Visual Genome (VG) \cite{goyal2017making} datasets contains 5.4M region descriptions for 108K images and 42 for each image on average. Each description phrase varies from 1 to 16 words. The dataset is densely annotated with scene graphs containing bounding boxes, classifications and attributes of main objects, and the relationships among different instances. Totally, it contains 3.8 million object instances, 2.8 million attributes and 23 million relationships. The R-CNN feature pre-trained \cite{anderson2018bottom} on VG dataset is employed in our experiment.

\begin{table*}
\begin{center}
    \small
    \begin{tabular}{|l|c|c|c|c|c|c|c|c|c|c|c|c|}
    \hline
    ~              & \multicolumn{6}{c|}{Cross-entropy loss} & \multicolumn{6}{c|}{CIDEr optimization training}    \\ \hline
    Models         & B-1                & B-4  & M    & R    & C                           & S    & B-1  & B-4  & M    & R    & C     & S    \\ \hline
    Test-guide \cite{Mun2016TextguidedAM}     & 74.9               & 32.6 & 25.7 & -    & 102.4                       & -    & -    & -    & -    & -    & -     & -    \\
    SCST \cite{Rennie_2017_CVPR}          & -                  & 30.0 & 25.9 & 53.4 & 99.4                        & -    & -    & 34.2 & 26.7 & 55.7 & 114.0 & -    \\
    StackCap  \cite{gu2018look}     & 76.2               & 35.2 & 26.5 & -    & 109.1                       & -    & 78.5 & 36.1 & 27.4 & -    & 120.4 & -    \\
    CAVP  \cite{liu2018context}         & -                  & -    & -    & -    & -                           & -    & -    & \textbf{38.6} & 28.3 & \textbf{58.5} & 126.3 & 21.6 \\ \hline
    Up-Down \cite{anderson2018bottom}   & \textbf{77.2}               & 36.2 & 27.0 & 56.4 & 113.5                       & 20.3 & 79.8 & 36.3 & 27.7 & 56.9 & 120.1 & 21.4 \\
    Ours:SG        & 77.1               & 36.3 & 27.8 & 56.8 & 115.3                       & 21.0 & 80.2    & 38.3    & 28.5    & 58.3    & 127.3    & 22.0    \\
    Ours:SG+RWS    & 77.1               & 36.6 & 28.0 & 56.9 & 116.9                       & 21.3 & 80.3 & 38.3 & 28.5 & 58.3 & 128.3 & 22.2 \\
    Ours:SG+RWS+WR & 77.1               & \textbf{36.6} & \textbf{28.0} & \textbf{56.9} & \textbf{116.9}                       & \textbf{21.3} & \textbf{80.3} & 38.5 & \textbf{28.7} & 58.4 & \textbf{129.1} & \textbf{22.4} \\ \hline
    \end{tabular}
\caption{Experiment of our proposed recall mechanism on the MSCOCO Karpathy test split with both cross-entropy loss and CIDEr optimization. We implement our proposed methods: semantic guide (SG), recalled-word slot (RWS) and recalled-word reward (WR) on the baseline model Up-Down. Test results show that our proposed methods have obvious improvement over our baseline. B-1 / B-4 / M / R / C / S refers to BLEU1/ BLEU4 / METEOR / ROUGE-L / CIDEr / SPICE scores.}
\label{tab:1}
\end{center}
\end{table*}

\subsection{Implementation Details}
For a more fair and convincing comparison, we use the R-CNN feature in Up-Down \cite{anderson2018bottom} as our image feature. For each image, 10 to 100 ROI (region of interest) pooling vectors are preserved. Each vector size is 2048. We use this R-CNN feature both in text-retrieval module and captioning model. Following previous methods in captioning, we do not use pretrained word embeddings, and all word embeddings are trained from scratch.

In our text-retrieval module, we adopt the mean pooling vector of R-CNN feature, and project it to a hidden vector of size 1024. Then we apply Bi-LSTM with size 512 to embed text. The word embedding size is also 512, so the output vector of Bi-LSTM is 1024, which matchs the hidden vector size of image. During training, the batch size is set to 128, and the margin $\alpha$ is set to 0.2. The learning rate is set to 5e-4 and decay by a factor 0.8 for every 3 epochs.

In captioning model, our base model is Up-Down model, so we use the same hyper-parameters as Up-Down model. The hidden units of LSTM1 and LSTM2 are both 1024, and the size of word embedding is also 1024. We adopt Adam optimizer with the learning rate set as 5e-4 and decay also by a factor 0.8 for every 3 epochs, and the batch size is set to 64. For CIDEr optimization training, we initialize the learning rate as 5e-5, decaying by a factor 0.1 for every 50 epochs. We choose the best model from cross-entropy training for the following CIDEr optimization training. GeForce GTX 1080Ti is the GPU we employed in all experiments.

\subsection{Performance of text-retrieval model}
The performance of text-retrieval model can directly affect the relevance between the recalled words and image. So we evaluate its performance on the validation set of MSCOCO Karpathy splits, and the result is reported in Table \ref{tab:3}. We assume that our text-retrieval model does not achieve state-of-the-art as in \cite{lee2018stacked}, but it is well qualified to retrieve  sufficient and relevant words for images.
Our corpus is collected from all the captions of MSCOCO Karpathy train splits. For determining an appropriate number of captions to be retrieved, we respectively retrieve top 1, 5 and 15 related captions for each image, and test the performance on cross-entropy loss. Table \ref{tab:4} shows that top 5 captions retrieved for each image is a better choice than 1 or 15. It is necessary to emphasize that we avoid retrieving the ground truth caption for images, and all the retrieved captions only come from the train splits of MSCOCO Karpathy. In the following experiments, top 5 captions retrieved are used to construct recalled words for each image.

\begin{table}[h]
\begin{center}
    \small
    \begin{tabular}{|l|c|c|c|c|c|}
    \hline
    Models & B-4  & M    & R    & C & S\\ \hline
    att2in \cite{Rennie_2017_CVPR}  & 36.1 &27.2&56.9 & 119.1&20.8\\
    att2in+SG+RWS+WR & \textbf{36.7}&\textbf{27.8} & \textbf{57.4} & \textbf{122.0}&\textbf{21.4}\\ \hline      
    att2all \cite{Rennie_2017_CVPR}  & 36.3 &27.5&57.2 & 121.7&21.1\\
    att2all+SG+RWS+WR  & \textbf{37.1}&\textbf{28.0}& \textbf{57.8} & \textbf{125.0}&\textbf{21.7}\\ \hline  
    stackcap \cite{gu2018look} & 36.6 &27.6& 57.3 & 121.1&21.0\\
    stackcap+SG+RWS+WR & \textbf{37.8} &\textbf{28.3}& \textbf{58.0} & \textbf{126.4}&\textbf{21.9}\\ \hline
    \end{tabular}
\caption{Performance of our proposed methods over other state-of-the-art models after cider optimization training.}
\label{tab:2}
\end{center}
\end{table}

\begin{table}[h]
\begin{center}
    \small 
    \begin{tabular}{|c|c|c|c|c|c|}
    \hline
    ~               & R@1  & R@5  & R@10 & Mean r \\ \hline
    Text-retrieval  & 36.2 & 69.0 & 81.7 & 7.7    \\ \hline
    Image-retrieval & 39.6 & 72.3 & 83.7  & 11.0   \\ \hline
    \end{tabular}
\caption{Performance of text-retrieval model on MSCOCO Karpathy validation set.}
\label{tab:3}
\end{center}
\end{table}

\begin{table}[h]
\begin{center}
    \small
    \begin{tabular}{|c|c|c|c|c|c|c|}
    \hline
    ~     & \multicolumn{6}{c|}{Cross-Entropy Loss} \\ \hline
    Top $K$ & B-1 & B-4  & M    & R    & C     & S    \\ \hline
    $K$=1   & 77.1 & 36.3 & 27.8 & 56.9 & 115.8 & 21.2 \\ \hline
    $K$=5   & 77.1 & 36.5 & 28.0 & 57.0 & \textbf{116.7} & 21.3 \\ \hline
    $K$=15  & 77.0 & 36.3 & 27.9 & 56.6 & 115.6 & 21.0 \\ \hline
    \end{tabular}
\caption{Experiments on choice of $K$, the number of captions retrieved for each image. B-1 / B-4 / M / R / C / S refers to BLEU1/ BLEU4 / METEOR / ROUGE-L / CIDEr / SPICE scores. Experiments are conducted on MSCOCO Karpathy validation set.}
\label{tab:4}
\end{center}
\end{table}

\begin{table}
\begin{center}
    \small
    \begin{tabular}{|c|c|c|c|c|c|c|c|}
    \hline
    ~   & \multicolumn{6}{c|}{Cider optimization training} \\ \hline
    $\lambda$   & B-1                         & B-4 & M & R & C & S  \\ \hline
    0.1 & 80.2                           & 38.2   & 28.5 & 58.2 & 127.5 & 22.3  \\ \hline
    0.3 & 80.2                           & 38.4   & 28.6 & 58.1 & 128.1 & 22.3 \\ \hline
    0.5 & 80.3                           & 38.4   & 28.6 & 58.3 &\textbf{128.9} & 22.4  \\ \hline
    0.7 & 80.3                           & 38.1   & 28.6 & 58.2 & 127.8 & 22.2 \\ \hline
    0.9 & 80.1                           & 38.0   & 28.4 & 58.1 & 127.3 & 22.2  \\ \hline
    \end{tabular}
\caption{Experiments on choice of $\lambda$, the trade-off parameter in CIDEr optimization. B-1 / B-4 / M / R / C / S refers to BLEU1/ BLEU4 / METEOR / ROUGE-L / CIDEr / SPICE scores. Experiments are conducted on MSCOCO Karpathy validation set}
\label{tab:5}
\end{center}
\end{table}

\begin{figure*}[t]
    \centering
    \includegraphics[width=0.95\textwidth]{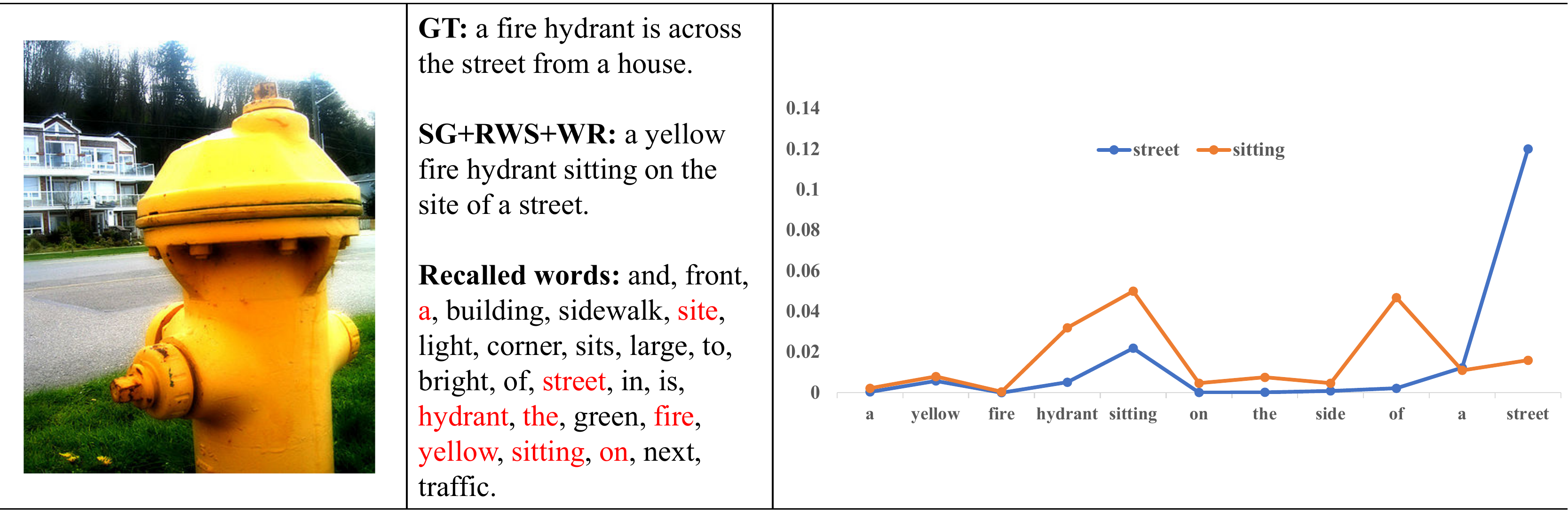}
    \caption{The left part:  recalled words and caption generated by SG+RWS+WR. The right part: visualization of the weights in recalled-word attention}
    \label{fig:3}
\end{figure*}

\begin{figure}[h]
    \centering
    \includegraphics[width=0.95\columnwidth]{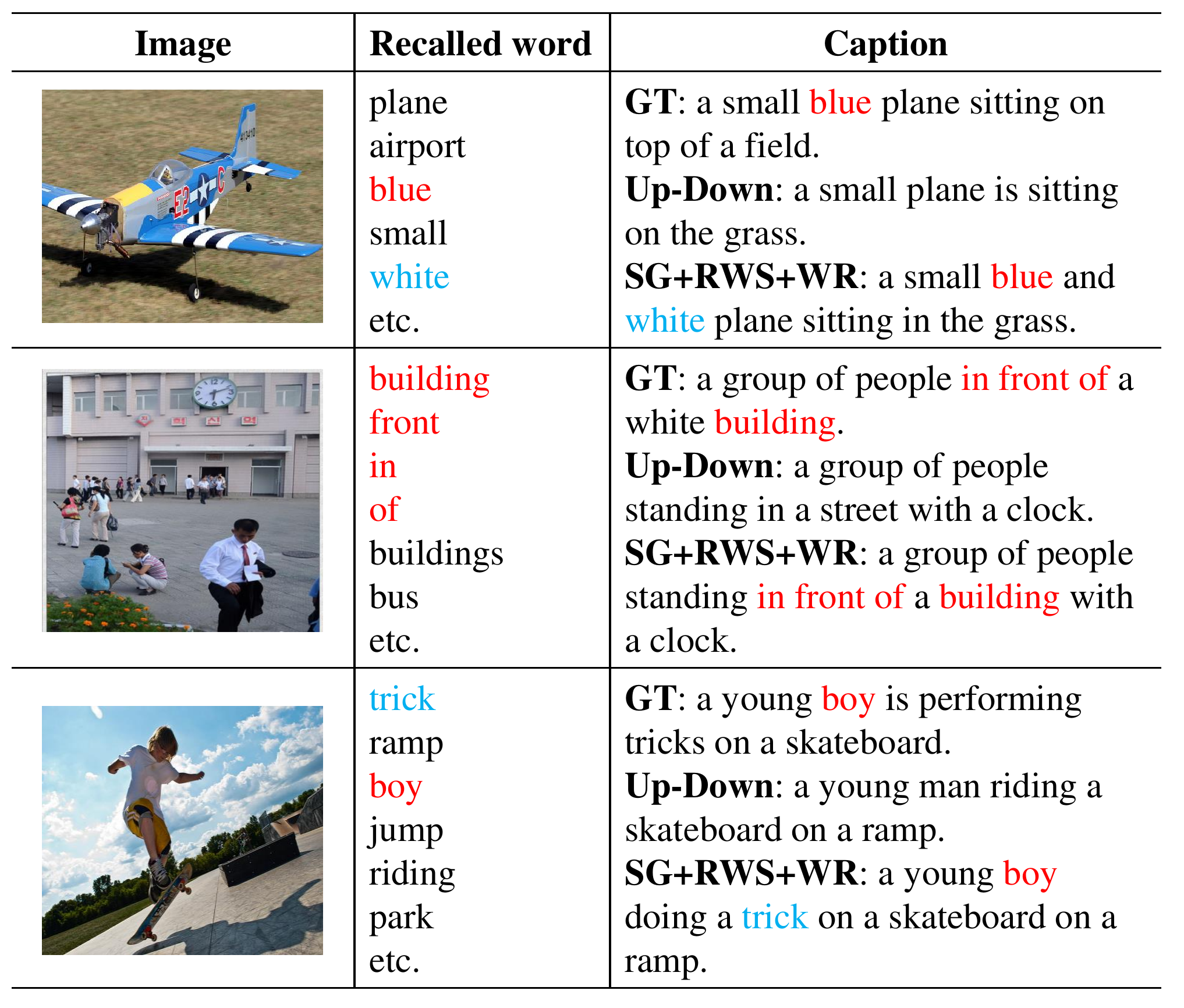}
    \caption{Recalled word and caption generation results on MS
COCO Karpathy test split 
and the output sentences are generated by 1) Ground Truth (GT):
one ground truth caption, 2)Up-Down model and 3) our SG+RWS+WR model. }
    \label{fig:4}
\end{figure}

\subsection{Captioning performance}
\subsubsection*{Selection of $\lambda$}
$\lambda$ in Eq.(\ref{equ:lam}) is the trade-off parameter in CIDEr optimization training, which balances the reward between $r(w_{1:T}^{\hat{s}})$ and $r(W_{r})$ in loss function. Thus we set different $\lambda$ values from 0 to 1 to conduct a model selection. Experiments for $\lambda$ are based on the best performance model ($K$=5) in Table \ref{tab:4}. The experiment result is reported in Table \ref{tab:5}, which shows that $\lambda = 0.5$ has the best performance out of others. As a result, in the following experiments, we set $\lambda = 0.5$.

\subsubsection*{Evaluation of proposed methods}
From above, we have proposed semantic guide (SG), recalled-word slot (RWS) in captioning model and recalled-word reward (WR) in CIDEr optimization. Then we test their performances on MSCOCO Karpathy test split. Beam search with beam size 2 is employed to generate captions. The performances of proposed methods are shown in Table \ref{tab:1}, and our baseline is Up-Down model in \cite{anderson2018bottom}. Focusing on the CIDEr score at Table \ref{tab:1}, semantic guide (SG) has improved 1.5\%  with cross-entropy loss, and 6.0\% on CIDER optimization. Semantic guide with recalled-word slot (SG+RWS) has improved 3.0\% on cross-entropy loss, and 6.8\% on CIDEr optimization. Our best model (SG+RWS+WR) has obtained 7.5\% improvement on CIDEr optimization, and it has obtained BLEU4 / CIDEr / SPICE scores of 36.6 / 116.9 / 21.3 with cross-entropy loss and 38.5 / 129.1/ 22.4 with CIDEr optimization. In addition, compared with other state-of-the-art models, like SCST \cite{Rennie_2017_CVPR}, StacKCap \cite{gu2018stack}, and CAVP \cite{liu2018context}, our results still outperform theirs. The results of comparison demonstrate the effectiveness of our proposed methods, especially on those more convincing evaluation metrics such as CIDEr, SPICE.

To prove the effectiveness and generality of our proposed methods, we have also implemented our proposed methods over other state-of-the-art models: att2in\cite{Rennie_2017_CVPR}, att2all\cite{Rennie_2017_CVPR} and stackcap \cite{gu2018look}. We do the comparative experiments over these three models. As is shown in Table \ref{tab:2}, the results indicate that our proposed methods have a wide range of applicability to many state-of-art models. To be detailed, we have average 2.3\% improvement on att2in, 2.1\% improvement on att2all, and 3.3\% improvement on stackcap. We have conducted the MSCOCO online evaluation and achieved promising results (called “caption-recall”, reported at 19 Oct 2019), which also surpass the online results of our baseline model Up-Down.

\subsubsection*{Qualitative Analysis}
To help qualitatively evaluate the effectiveness of our recall mechanism, Figure \ref{fig:3} and Figure \ref{fig:4} show some examples generated by our recall mechanism. As shown in Figure \ref{fig:4}, we can observe that recalled words whose font in red color, like: ``blue", ``front", ``building" and ``boy", are generated in captions by SG+RWS+WR, and those words are also in ground truth, but not in the caption generated by the base model Up-Down. This illustrates that our recall mechanism make the generated sentence closer to ground truth caption. Moreover, there are some recalled words in blue color that are not in ground truth caption, but they are highly consistent with the image. In Figure \ref{fig:3}, we present all the recalled words for an image. In this example, each word in generated caption can be found in recalled words. This proves the high correlation between generated caption and recalled words. Recalled-word slot and semantic guide are highly dependant on the weights in recalled-word attention, so we also visualize the weights of ``sitting" and ``street" at each generation step. We can observe that two words weight attain the max when they are generated.

\section{Conclusion}
In this paper, we introduce a novel recall mechanism to the captioning model. In our recall mechanism, a recall unit is designed to retrieve recalled words for image, and semantic guide and recalled-word slot are proposed to make full use of recalled words. A recalled-word reward is introduced to boost CIDEr optimization. The experiments prove that our recall mechanism can effectively employ recalled information to improve the quality of generated caption.

It needs to be emphasized that our proposed methods can be applied to any captioning model. Meanwhile, using training data both for model training and retrieving recalled words, can be instructive to other researches of captioning.

\section{Acknowledgement}
This paper is supported by NSFC (No.61772330, 61533012, 61876109), the Basic Research Project of Innovation Action Plan (16JC1402800), the advanced research project (No.61403120201), Shanghai authentication key Lab. (2017XCWZK01),Technology Committee the interdisciplinary Program of Shanghai Jiao Tong University (YG2015MS43), and AI-Lab VS Team, Bytedance.

\bibliography{3148.aaai_20.bib}
\bibliographystyle{aaai}

\end{document}